\documentclass[journal]{IEEEtran}

\usepackage{lipsum}
\usepackage{graphicx}
\usepackage{cite}
\usepackage{amsmath,amssymb,amsfonts}
\newtheorem{remark}{Remark}  
\usepackage[linesnumbered,ruled]{algorithm2e}
\usepackage[usenames,dvipsnames,svgnames]{xcolor}
\usepackage{adjustbox}
\usepackage{covington}
\SetKw{KwBy}{by}
\usepackage{caption} 
\usepackage{multirow}
\usepackage{xcolor}
\usepackage{comment}
\usepackage{float}
\usepackage{etoolbox}

\usepackage{scalerel}
\usepackage{tikz}
\usetikzlibrary{svg.path}

\definecolor{orcidlogocol}{HTML}{A6CE39}
\tikzset{
  orcidlogo/.pic={
    \fill[orcidlogocol] svg{M256,128c0,70.7-57.3,128-128,128C57.3,256,0,198.7,0,128C0,57.3,57.3,0,128,0C198.7,0,256,57.3,256,128z};
    \fill[white] svg{M86.3,186.2H70.9V79.1h15.4v48.4V186.2z}
                 svg{M108.9,79.1h41.6c39.6,0,57,28.3,57,53.6c0,27.5-21.5,53.6-56.8,53.6h-41.8V79.1z M124.3,172.4h24.5c34.9,0,42.9-26.5,42.9-39.7c0-21.5-13.7-39.7-43.7-39.7h-23.7V172.4z}
                 svg{M88.7,56.8c0,5.5-4.5,10.1-10.1,10.1c-5.6,0-10.1-4.6-10.1-10.1c0-5.6,4.5-10.1,10.1-10.1C84.2,46.7,88.7,51.3,88.7,56.8z};
  }
}

\newcommand\orcidicon[1]{\href{https://orcid.org/#1}{\mbox{\scalerel*{
\begin{tikzpicture}[yscale=-1,transform shape]
\pic{orcidlogo};
\end{tikzpicture}
}{|}}}}

\usepackage[hidelinks]{hyperref}

\ifCLASSINFOpdf

\else

\fi

\begin{document}

\title{Receding Horizon Optimization with PPUM: An Approach for Autonomous Robot Path Planning in Uncertain Environments
}

\author{

Zijian Ge \orcidicon{0000-0002-9620-3843}\,,
Jingjing Jiang \orcidicon{0000-0001-7754-9147}\,, \IEEEmembership{ Member, IEEE},
Matthew Coombes \orcidicon{0000-0002-4421-9464}\,,
Liang Sun \orcidicon{0000-0003-3856-1383}\,, \IEEEmembership{Member, IEEE},
\thanks{Zijian Ge, Jingjing Jiang and Matthew Coombes are with the Department of Aeronautical and Automotive Engineering, Loughborough University, Leicester LE11 3TU, UK, (e-mail:z.ge@lboro.ac.uk; j.jiang2@lboro.ac.uk; m.j.coombes@lboro.ac.uk).} 
\thanks{ Liang Sun is with the School of Intelligence Science and Technology, University of Science and Technology Beijing,Beijing 100081, China, (e-mail:liangsun@ustb.edu.cn).} 
}

\markboth{ }%
{Shell \MakeLowercase{\textit{et al.}}: Bare Demo of IEEEtran.cls for IEEE Journals}

\maketitle

\begin{abstract}

The ability to understand spatial-temporal patterns for crowds of people is crucial for achieving long-term autonomy of mobile robots deployed in human environments. However, traditional historical data-driven memory models are inadequate for handling anomalies, resulting in poor reasoning by robot in estimating the crowd spatial distribution. In this article, a Receding Horizon Optimization (RHO) formulation is proposed that incorporates a Probability-related Partially Updated Memory (PPUM) for robot path planning in crowded environments with uncertainties. 
The PPUM acts as a memory layer that combines real-time sensor observations with historical knowledge using a weighted evidence fusion theory to improve robot's adaptivity to the dynamic environments. RHO then utilizes the PPUM as a informed knowledge to generate a path that minimizes the likelihood of encountering dense crowds while reducing the cost of local motion planning. The proposed approach provides an innovative solution to the problem of robot's long-term safe interaction with human in uncertain crowded environments. In simulation, the results demonstrate the superior performance of our approach compared to benchmark methods in terms of crowd distribution estimation accuracy, adaptability to anomalies and path planning efficiency.

\end{abstract}

\def\abstractname{Note to Practitioners}
\begin{abstract}
For practitioners, our work motivated by the need of long-term deployment of mobile robot in human environments, it offers adaptive path planning that integrates sensor data and historical knowledge, enhancing robot's adaptability in dynamic settings. RHO employs PPUM to create optimal paths, ensuring efficient and safe navigation through crowded spaces. Importantly, our method addresses the challenges of crowded spaces, considering both short-term sensor data and long-term crowd patterns. However, implementing this approach requires a long-period data collection, careful parameter tuning and real-world testing, considering data fusion and optimization techniques. By adopting our method, practitioners can empower robots to proficiently navigate complex settings characterized by the coexistence of robots and humans, making them suitable for diverse applications involving crowd interactions.
\end{abstract}


%
\IEEEpeerreviewmaketitle

\section{Introduction}
%
%
%
%
\IEEEPARstart{T}{he} deployment of mobile robots in human environments has become increasingly common in recent years. Path planning is a critical aspect of mobile robot navigation in both industrial and service applications. 
Typically, a robot path planning algorithm determines a collision-free path from start location to goal location while optimizing some specific objectives, such as congestion cost\cite{lin2020spatiotemporal}, human comfort\cite{morales2013human}, or energy consumption\cite{tokekar2014energy}. Due to the safe interaction and robust long-term autonomy of the robots. The awareness of the crowd spatial-temporal distribution can be a significant requirement for those robots deployed in human environments. The presence of crowded spaces can significantly increase the computational burden on the local planner and result in longer travel times or even crashes with pedestrians.

Consider a scenario in which a robot has been deployed in a space where robots and humans coexist for an extended period. Over time,  it collects historical data regarding the spatial-temporal patterns of crowds, which depicts how crowds move and interact over time. For example, these patterns might include how pedestrians gather in certain areas during peak hours, how they disperse during quieter periods, or how they navigate around obstacles or each other. Given the prior knowledge, the goal for the robot is to plan a global reference path that avoids potentially crowded areas. However, in practice, the environment is uncertain and contains anomalies, which can refer to an unexpected or unusual behavior or patterns exhibited by a group of people in a particular location or situation. Anomalies can be caused by various factors, such as external events or holidays, leading to unexpected changes in crowd density or movements that deviate from typical behavior. The ability to detect and respond to crowd anomalies is also essential for ensuring the safety and efficiency of mobile robots operating in human environments, as it enables them to avoid crowded areas and navigate through complex and dynamic environments with less risks, as well as help robots work in a socially acceptable manner to avoid causing discomfort or annoyance to people.

\begin{figure}[!t]
	\hspace*{-0.0cm}
	\vspace*{0.2cm}
	\includegraphics[width=0.5\textwidth]{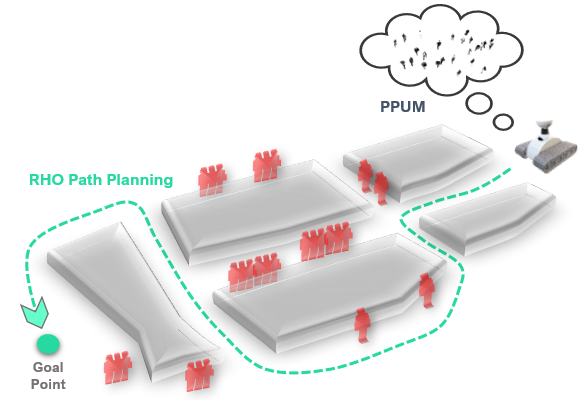}
	\caption{Illustration demonstrating how the PPUM assists in RHO path planning to find a path that avoids crowded areas (Pedestrians are highlighted in red).}
	\label{fig:illustration}
\end{figure}

The estimation of crowd spatial-temporal distribution for the long-term autonomy of robots has gained considerable interest in the research community. Several algorithms, such as the spectra-based model in \cite{Krajnik2014}\cite{Fentanes2017} and convolutional recurrent neural network model for crowd's macroscopic properties prediction \cite{Kiss2021}, have been proposed to enhance robots' reasoning in dynamic environments. 
In fact, it is important to note that the spatial-temporal pattern prediction is not limited to crowds and is extensively investigated in the general domain of Intelligent Transportation Systems for autonomous vehicles. 
Deep learning-based approaches \cite{miglani2019deep} \cite{Luo2019} \cite{Chen2020a} for traffic flow prediction, and hazard-based models for traffic incident clearance time prediction \cite{hojati2013hazard} \cite{nam2000exploratory}\cite{junhua2013estimating} are examples of such methods. 
However, those data-driven prediction methods often neglect outliers, which can lead to overfitting problems in prediction. 
Consequently  these methods may struggle to adapt to uncertain environments in the present of anomalies. This lack of adaptability can lead to misguidance in robot's route selection, potentially causing navigation delays or even hazards\cite{gecongestion}. 
Therefore, it is important to address the challenges of outliers and anomalies in the robots' reasoning system to ensure the robustness and reliability of autonomous systems in dynamic and unpredictable scenarios.

Another active research area focuses on path planning strategies to avoid crowded areas, often referred to as motion planning. Works such as \cite{chen2019crowd} and \cite{nishimura2020l2b} address safe and efficient robot navigation while considering social expectations in human environments \cite{gao2022evaluation}. For example, in the study conducted by \cite{Vega2019}, the authors propose a crowd clustering approach utilizing a personal space model to identify forbidden regions where humans engage in specific activities. The objective is to enable robots to exhibit socially acceptable behaviors during navigation. However, these local planning methods only react to crowds that are directly observed by sensors and lack the capability to anticipate crowd spatial distribution in advance. In highly crowded areas, robots may experience local freezing, impeding their movement. This issue could be mitigated if robots possess prior knowledge of the crowded areas before initiating their navigation. Other global path planning methods with prior information, such as the graph-based path searching method A* developed in \cite{Zhang2020a,gecongestion}, where the congestion cost is integrated into the searching step, and the artificial potential field-based method in \cite{basu2003routing}, where crowded areas are formulated as repulsive potential fields, primarily focus on avoiding congestion based on known information at the global level. However, these traditional heuristic-based planning methods may struggle to provide an optimal solution for crowd scenarios due to their simplistic nature. An optimal solution in crowded settings encompasses more than just avoiding collisions. It entails finding a path that not only avoids obstacles but also takes into account the dynamic nature of crowds. Such a solution should prioritize both efficiency and safety.

To address these challenges, a bio-inspired memory model called PPUM is developed to enhance the cognition and social awareness of robots in understanding human macroscopic behaviors. The PPUM consists of two memory layers: the Off-line Memory (OLM) layer and the Working Memory (WM) layer. The OLM layer utilizes historical prior knowledge to represent the dynamic probability distribution of the crowd, while the WM layer is continuously updated in real-time through sensor measurements, providing likelihood evidence of the current crowd dynamics. By fusing the information from both memory layers using a Weighted D-S evidence combination rule, a comprehensive understanding of the crowd's spatial-temporal patterns is achieved. This knowledge is then leveraged in a novel Receding Horizon Optimization (RHO)-based global path search strategy, which provides an approximation of a globally optimized solution for the path planning problem. By incorporating the PPUM, the path planning algorithm can effectively generate paths that minimize the likelihood of encountering dense crowds, even in uncertain environments. An illustration of how the proposed method work in the human environment can bee seen in Fig. \ref{fig:illustration}.

This study represents a significant advancement over prior research \cite{gecongestion}, where the concept of path planning utilizing multi-memory layers was initially proposed. The key contributions of this work are outlined below.

\begin{figure*}[t]
    \centering
    \includegraphics[width=0.8\textwidth]{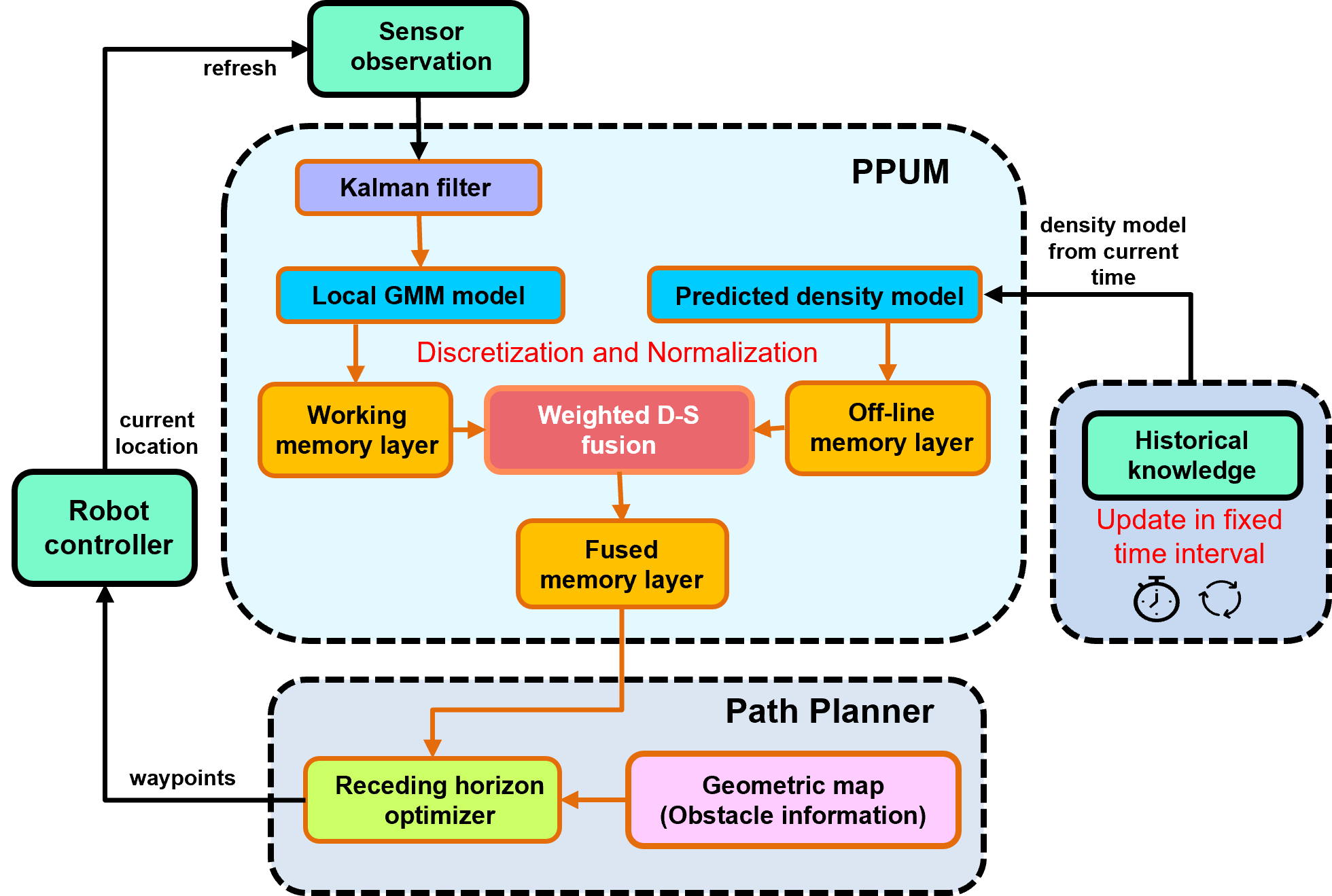}
    \caption{System architecture of the proposed framework.}
    \label{fig:system}
\end{figure*}

\begin{itemize}
    \item  An innovative multi-layer crowd congestion probability density map is proposed, utilizing the Gaussian Mixture Model as bio-inspired memory layers to accurately estimate the spatial distribution of crowds.
    

    \item We introduce a novel method called Probability-Related Partially Updated Memory (PPUM), which builds upon the Partially Updated Memory (PUM) concept. PPUM utilizes the weighted DS evidence theory to enhance a robot's reasoning in uncertain and dynamic environments.

    \item A receding horizon optimization formulation is developed for congestion-aware path planning that accommodates the PPUM of mobile robots.

\end{itemize}

The remainder of the paper is organized as follows: The construction of the PPUM is introduced in Section \ref{secppum} and RHO for path planing is presented in Section \ref{secrho}. 
The simulation results and comparative case studies are described and discussed in Section \ref{secsim}. The paper is finally concluded in Section \ref{seccon}.



\section{Construction of the PPUM}\label{secppum}
Human cognition encompasses various traits, such as short and long-term memory, aiding in categorization, conceptualization, reasoning, planning, problem-solving, learning, and creativity, these abilities collectively reflect the complex cognitive processes involved in human intelligence and adaptive behavior \cite{dodd2005role}\cite{Zou2016}. Drawing inspiration from the human cognitive system and Bayesian theory, we propose the PPUM. Leveraging sensor observations as real-time evidence and off-line historical data as prior evidence, the PPUM derives posterior evidence. The overview of memory working process can be viewed in Fig.~\ref{fig:system}, and this section aims to explain the generation of all the memory layers and the proposed method for memory fusion.

\subsection{Working Memory layer}
WM of the robot in this case is also defined as the sensor memory which contains an  active memory responsible fo temporarily storing the current observation while in a task \cite{dodd2005role}\cite{persiani2018working}.  
It is assumed that the observations of the robots are obtained from widely used Lidar sensors and follow a Gaussian noise distribution. Subsequently, the local crowd probability density distribution  can be derived from a sensor filter, which can be taken as the WM layer in our work.

\subsubsection{Crowd's probability density map generation}
Since sensor noise assumed to be normal white noise, then a Kalman filter\cite{welch1995introduction} can be employed due to its efficient computation on handling linear system,  to track and generate the probability density distribution of the moving pedestrians.

To elaborate, the process begins with the sensor tracking each pedestrian's movements independently. The Kalman filter then refines these individual estimates. Ultimately, these enhanced estimates are amalgamated to form a statistical model known as a Gaussian Mixture Model (GMM), which characterizes the broader behavior of the crowd. Let \( n_p \) represents the number of pedestrians detected by the sensor. With posterior covariance  \(P_{k}\), and state \( \hat{x}_{k} \) at current time step, the spatial normal distribution regarding to the $i^{th}$ pedestrian's position can be generated, which can be written as the following density function: 
\begin{equation*} \label{eq10}
  p(x|\hat{x}_{k}^i,P_{k}^i)= {\frac{ exp\{ -\frac{1}{2} (x-\hat{x}_{k}^i)^{'} (P_{k}^i)^{-1}  (x-\hat{x}_{k}^i) \}}{ (2\pi)^{1/2} |P_{k}^i|^{1/2} } }
\end{equation*}
Subsequently, given the {state} of each pedestrian:
\begin{equation*} \label{eq11}
  \lambda=\{\hat{x}_{k}^i,P_{k}^i,w_i \},  \quad i=1,...,n_p,
\end{equation*}
where $w_i$ represents {the weight of each component}.

The observed local crowd spatial distribution can be described by follows using GMM:
\begin{equation} \label{eq12}
  p(x|\lambda)= \sum_{i=1}^{n_p} w_i g(x|\hat{x}_{k}^i,P_{k}^i) 
\end{equation}
The mixture weight of each component, denoted by \( w_i \), is subject to the constraint  \(  \sum_{i=1}^{n_p} w_i = 1  \) in GMM. In our case, the weights are equally distributed and assigned the value \( w_i = \frac{1}{n_p} \). Prior to using the model in equation (\ref{eq12}) as the WM layer, it requires a processing step involving discretization and normalization, which is further explained in Remark \ref{rm1}.

\subsection{Off-line Memory layer}
The off-line memory (OLM) of the robot stores the data, including historical sensor readings, maps, learned models, and past experiences. In this work, it is considered a representation of the typical motion patterns exhibited by people, as studies have shown that regular crowd spatial-temporal patterns are predictable \cite{Fentanes2015}. 
By utilizing the OLM, the robot can obtain predicted density information based on the specific time and location. 
To collect the dataset, a long-term deployment of fixed lidar sensors in a corridor can be employed, enabling the tracking of crowd spatial-temporal behaviors over extended duration  (\emph{i.e.}, months and even years). Detailed information on this technique can be found in \cite{B2018}.

\subsubsection{Crowd's probability density map prediction}It is assumed that historical data set is available and ready to use. The crowd density prediction can be conducted by  Warped Hypertime Representation \cite{Vintr2019}\cite{Krajnik2019}, which is an extended method based on the Frequency Map Enhancement in \cite{Fentanes2017}. This model introduces a function \( \rho(x_o,t_o) \) that captures the frequency of a specific vector \( x_o \) at a given time \( t_o \). To enhance its representation and analysis, the function is transformed using several clusters within the “Wrapped Hypertime-Space”, and the process for the prediction is detailed in \cite{Vintr2019}.

\begin{remark}\label{rm1}
Let us divide the working space of the robot \(M\) by \(n \times n\) cells with space length \(l\). Since both memory layers are exhibited as the GMM model, the volume of each cell \((i,j) \in M\) satisfy \( \sum_{i,j \in M} V_{ij} = 1 \). 
To make the GMM model a probability distribution which satisfy the constraint \( \sum_{i,j \in M} p(x_{ij}|\lambda)= 1 \), the following normalization steps are conducted:
\begin{equation} \label{eq14}
	\hat{p}(x_{ij}|\lambda) = \frac{p(x_{ij}|\lambda)}{\sum_{i,j \in M} p(x_{ij}|\lambda)} \times \frac{1}{a} \times a, \quad a= (\frac{l}{n})^2
\end{equation}
where, \(a\) is the area of each cell, and normalized probability of each cell is obtained by \(\hat{p}(x_{ij}|\lambda)\). The normalization of probabilities across the entire space ensures that the probabilities accurately reflect the relative likelihoods of each cell being crowded, and is necessary for the subsequent evidence fusion process, which will be discussed in detail in the following section. 
\end{remark}

\subsection{Fused memory layer}
The process of memory fusion can be conceptualized as a trust assignment between two weighted evidence sources. In this context, the evidence sources represent different memory layers that contribute to the overall understanding and perception of the system. The fusion process involves combining and weighing the evidence from each source based on their reliability and credibility. This trust assignment ensures that the resulting Fused Memory (FM) accurately represents the collective knowledge and information from both sources, leading to a more robust and comprehensive representation of the crowd density distribution.

\subsubsection{Dempster-Shafer Theory}
The Dempster-Shafer (D-S) theory of evidence is first introduced and finalized in \cite{dempster2008upper}\cite{dempster1968generalization}\cite{shafer1976mathematical}, which is a data fusion technique for estimating the posterior probability, widely used in map fusion \cite{Li2014}\cite{Chen2018}\cite{Zhu2014} and fault diagnosis \cite{Chen2019}. 
Also, it is known as the generalisation of the Bayesian theory \cite{Beynon2000}. Compared to traditional Bayesian theory based methods, D-S theory offers a more flexible and adaptive approach to handling uncertainties, and allows for the representation of ignorance or lack of knowledge through the belief function.
For a position \((x,y)\) over the continue space \(U(x,y)\), is characterized by two states \(C\) and \(NC\), corresponding to the probabilities that the position is “Crowded” or “Not Crowded” . 
In D-S theory, the state is defined by the set of discernment by:
\begin{equation*} \label{eq15}
  \Lambda=2^{\theta}=\{ C,NC, \{C,NC\},\emptyset \}
\end{equation*}
where \(2^{\theta}\) is the size of the focal elements in discernment frame \(\Lambda\). 
The belief of a position in the working space \(U(x,y)\) is described by assigning basic probability \(m_{x,y}\), which is also defined as the mass function to a specific state. 
By assuming the probability of both states coexisting is zero, we have \( m_{x,y}(\{ C,NC \})=0\).
Hence,
the mass functions on the discernment frame 
must satisfy the following constraints:
\begin{equation*} \label{eq16}
	   m_{x,y}(\emptyset)=0, \quad 
     m_{x,y}(C) + m_{x,y}(NC) = 1
\end{equation*}

\subsubsection{Basic Probability Assignment}
By considering two information sources from the WM and OLM, the associated mass functions \( m_s \), \( m_f \) and their Basic Probability Assignment (BPA) \( m_s(C_{xy}) \), \( m_f(C_{xy}) \) for the state “Crowded”  \( C \) can be derived from their respective normalized GMM models, following equation (\ref{eq14}) as:
\begin{equation} \label{eq19}
	   m_s(C_{xy}) = \hat{p}_s(C_{xy}|\lambda_s), \quad 
     m_f(C_{xy}) = \hat{p}_f(C_{xy}|\lambda_f)
\end{equation}
where \( m_s(C_{xy}) \) for instance, indicates that the belief of the WM that the location \( (x,y) \) is crowded.

\subsubsection{Weighted memory combination rule}
Traditional D-S evidence theory treats each evidence equally, disregarding the information priorities. In evidence fusion, it is widely accepted that multiple evidence from different sources should be weighted based on their respective importance and reliability. In this work, the sensor observations from the robots are assigned higher credibility compared to the OLM. 
This parallels the principle in human recognition systems where what is “seen” carries more credibility than what is “remembered”. 
According to the Weighted Evidence Theory (WET) introduced in \cite{yongsheng2002study}, \( m_j^{'} \) denotes the weighted BPA of \( m_j \) given by the source \( S_j \), it is modified by weighted average \( \Bar{m} \) of all sources and the population variance, which can be written as follows:
\begin{equation*} \label{eq21}
	   m_j^{'}=m_j - \epsilon_j \delta_j
\end{equation*}
where \( \epsilon_j \) indicates the deviation of \( m_j \) from \(  \Bar{m} \), and the population variance is given by \( \delta_j \). 
They can be obtained by:
\begin{equation*} \label{eq22}
	   \epsilon_j =  m_j - \Bar{m}, \quad 
     \delta_j = \sum_{k=1}^{M} (\epsilon_{k,j})^2/M
\end{equation*}
where the size of the focal elements in source \( S_j \) is represented by \( M \). 
Since the deviation \( \epsilon_j \) and population variance \( \delta_j \) are typically small in magnitude, resulting in a negligible value of \( \epsilon_j \delta_j \). As a result, the modification of the evidence through this factor is not significant or noticeable. 

In the alternative method named Weighted Balance Evidence Theory (WBET)  proposed in \cite{guo2006weighted}, which balances the evidence by taking the weighted average of all the prior evidence before integrating them. Consider two information sources \( S_s \) and \( S_f \) from the WM and OLM respectively. Let \( w_s \) and \( w_f \) be the weight coefficients of their BPA: \( m_s \) and \( m_f \). According to the WBET, the following constraint should be satisfied:
\begin{equation*} \label{eq24}
\begin{split}
	  w_s+w_f=1, \quad w_s > w_f
\end{split}
\end{equation*}
where \( w_s > w_f\) indicates that the WM derived from sensor observation always has higher credibly. 
The weighting coefficients can be generally obtained through expertise or extensive statistical data \cite{guo2006weighted}. 
However, instead of arbitrarily selecting the weights, an alternative method is proposed for determining the weights based on the covariance of the Kalman filter, which is explained in Remark \ref{rm2}. 
The weighted average of all is calculated as:
\begin{equation*} \label{eq25}
	  \Bar{m}=w_s m_s + w_f m_f.
\end{equation*}
Subsequently, the balanced mass functions by weighted average \(  \Bar{m} \) and prior evidence can be defined as:
\begin{equation} \label{eq26}
    m_s^{'}=m_s, \quad m_f^{'}=2 \Bar{m}-m_f
\end{equation}
Finally, based on D-S evidence joint rules, the fused memory can be calculated by equation:
\begin{equation*} \label{eq27}
    m^{'}(C)= 
\begin{cases}
    \frac{\sum_{S_s \cap S_f = C}  m_s^{'}(S_s) \cdot m_f^{'}(S_f) }{1- \kappa},& C\neq \emptyset\\
    0,              & C = \emptyset
\end{cases}
\end{equation*}
where {\(  \kappa \)} is defined as the conflict degree of two sources \(  S_s \) and \(  S_f \) , obtained by:
\begin{equation*} \label{eq28}
 \kappa =\sum_{S_s \cap S_f = \emptyset} m_s^{'}(S_s) \cdot m_f^{'}(S_f)
\end{equation*}
\begin{remark}\label{rm2}
A proposed solution for the acquirement of weighting coefficients in this case is according to covariance of the Kalman filter. 
A lower covariance indicates a more precise and reliable sensor, leading to higher value assigned to \(  w_s \). 
On the other hand, a higher covariance implies more uncertainty and potential errors in the sensor measurements, which can negatively impact the tracking accuracy, leading to lower value assigned to \(  w_s \). 
In general an alternative expression for calculating weight coefficients of two memory sources can be rewritten as:
\begin{equation} \label{eq29}
w_s = \frac{e^{\Sigma \gamma}+1}{2}, \quad w_f=1-w_s.
\end{equation}
where \(  \Sigma \) denotes the average covariance derived from the GMM model \( \hat{p}_s(C_{xy}|\lambda_s) \), and \( \gamma \) is the scale factor. 
Note that equation. (\ref{eq29}), indicates \( w_s \in (0.5,1) \).
\end{remark}

\begin{figure}[!t]
    \centering
	\hspace*{-0.0cm}
	\vspace*{-0.0cm}
	\includegraphics[width=0.4\textwidth]{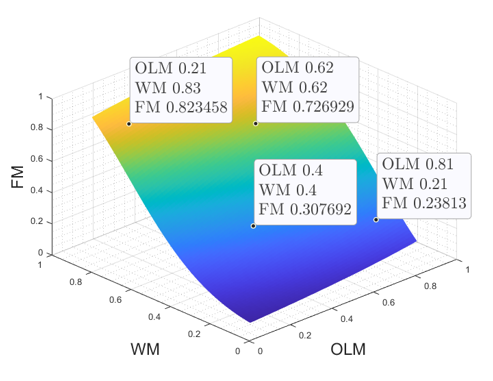}
	\caption{Fused probability outcomes of WM and OLM are depicted. In scenarios with substantial conflict between OLM and WM, the OLM's impact on the fused result diminishes, aligning it more closely with the WM's belief.}
	\label{fig:fusionsurface}
\end{figure}
\begin{figure}[!t]
	\hspace*{-0.0cm}
	\vspace*{-0.0cm}
	\includegraphics[width=0.45\textwidth]{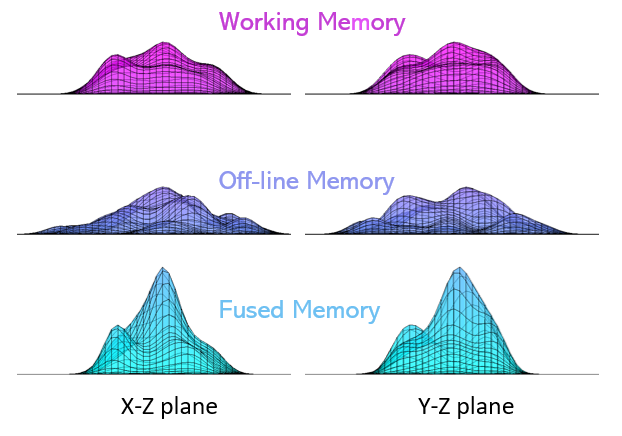}
	\caption{An example of the memory fusion result}
	\label{fig:fusinmemory}
\end{figure}

\begin{algorithm}[!t]
    \SetKwInOut{Input}{Input}
    \SetKwInOut{Output}{Output}
    \Input{BPA of WM $m_s$; 
    BPA of OLM $m_f$\;
    Covariance of Kalman filter $\Sigma$\;}
    \Output{FM: $m^{'}$;}

\While{$m_s$ is not empty}
      {
      
  \For{the overlaying area of $m_s$ and $m_f$ }
      {
        step 1: Memories get weighted by WBET:\\  
        $(m_{s}^{'},m_{f}^{'})$ $\leftarrow$ GetBalanced($m_{s},m_{f},\Sigma)$\;
        step 2: Memories get fused by D-S theory:\\
        $m^{'}$ $\leftarrow$ DSF($m_{s}^{'},m_{f}^{'}$)\;
      }
      }
\caption{PPUM}
\label{alg:ppum}
\end{algorithm}

In the approach of this work, the BPA of the WM and OLM are obtained directly using equation (\ref{eq19}). For the target state “Crowded” \(  C \) on \(  (x,y) \) , based on the weighted combination method in equation (\ref{eq26}), the update rule of the FM is: 
\begin{equation} \label{eq30}
m^{'}(C_{xy}) = { m_s^{'}(C_{xy}) \cdot m_f^{'}(C_{xy})  
  \over\displaystyle{
  1-(m_s^{'}(C_{xy}) \cdot m_f^{'}(NC_{xy})
  \atop\quad
  {}+ m_s^{'}(NC_{xy}) \cdot m_f^{'}(C_{xy}))}}
\end{equation}

The pseudo code of the update mechanism of PPUM can be viewed in Algorithm \ref{alg:ppum}, where \(  Getbalanced(m_s,m_f,\Sigma) \) calculates the weighted evidence according to equation (\ref{eq26}), and \(  DSF(m_s^{'},m_f^{'}) \) obtains BPA of the FM based on equation (\ref{eq30}). 
The probability fusion distribution can be seen in Fig.~\ref{fig:fusionsurface}, and Fig.~\ref{fig:fusinmemory} intuitively demonstrates the distribution of the memory after fusion.

\section{Receding Horizon Optimization for Path Planning}\label{secrho}
The objective of this section is to plan a global path for an autonomous robot in the presence of obstacles while minimizing the probability of navigating to crowded areas based on the information stored in PPUM. 
A novel Receding Horizon Optimization (RHO) formulation is proposed and adopted in this work (see Fig.~\ref{fig:rh_feature}). 
It provides more flexibility in incorporating multiple factors into the path planning process, resulting in more versatile and optimized path solutions.

\subsection{{Problem Formulation}}\label{costfunction}

\subsubsection{Cost Function}\label{sec:costfun}
As depicted in Fig.~\ref{fig:rh_feature}, the overall planned path is based on a series of sub-paths. 
To determine the $c^{th}$ sub-path $wp^c$,
the following optimization objective is given:
{
\begin{equation} \label{eq31}
wp^c =  {\rm argmin} \sum_{i=1} ^{n} ||wp^c_{i}-goal||  + p(d_{e2g}^c)\cdot \sum_{i=1} ^{n} P(wp^c_{i}) 
\end{equation}}
Each waypoint in \(  wp^c \), represented by \(  wp^c_i \in \mathbb{R}^{1 \times 2 } \) with \(  i \in \{1,2,\dots,n\} \), is evaluated using a combination of the distance to the goal and the function {\(  P(wp^c_i) \)} that calculates the probability of location \(  wp^c_i \) being crowded based on the FM generated from PPUM.
The dynamic weighting factor is calculated by the scale function \(  p(d_{e2g}^c)\), which is further explained in the corresponding Remark \ref{rm3}. 
By optimizing the sum of distance between goal location \(  goal \) and waypoint \(  wp^c_i \), {it guides the sub-path towards the direction of the goal. This ensures that the global reference path effectively reaches the destination.

\begin{figure}[!h]
    \centering
	\hspace*{-0.0cm}
	\vspace*{-0.0cm}
	\includegraphics[width=0.35\textwidth]{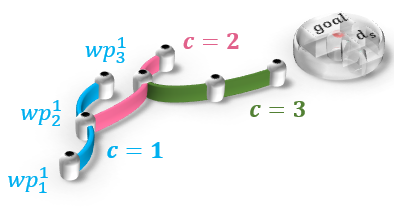}
	\caption{{Illustration of the iterations of the receding horizon-based path planning. \(d_s\) is the inflation radius of the goal, which is explained in detail in Section \ref{subsec ppm}.}}
	\label{fig:rh_feature}
\end{figure}

\begin{remark}\label{rm3}
The scale function \(p(d_{e2g}^c)\) for the probability cost is not a constant value as it varies according to the distance. 
As the end node of each sub-path approaches the goal, the distance cost decreases. 
To ensure that the density cost is in the same range as the distance cost, it needs to be adjusted accordingly, \emph{i.e.}, $p(d_{e2g}^c) = \alpha \cdot d_{e2g}^c$ where \(\alpha\) is the weighting factor and $d_{e2g}^c=||wp_{n}^c-goal|| $ denotes the distance between the end node of $c^{th}$ sub-path and the goal. 
\end{remark}

\subsubsection{Sub-path Generation}
Additional to the cost function given in Section \ref{sec:costfun}, the following constraints are also considered in the optimization problem:
{\begin{align}
         ||wp^c_{i+1}-wp^c_{i}|| \leq
         &\left\{
         \begin{array}{ c l }\label{eq33}
          d_{I},  & \quad \textrm{if } d_{e2g}^c > d_{r} \\
          \dfrac{d_{I}}{n},               & \quad \textrm{otherwise}
         \end{array}
          \right.\; 
         \\
         d_{wp}^c \leq
        &\left\{ 
         \begin{array}{ c l }\label{eq34}
          d_{l},  & \quad \textrm{if } d_{e2g}^c > d_{r} \\
          d_{e2g}^c,               & \quad \textrm{otherwise}
         \end{array}
          \right.\;   
\end{align}}
where \(  d_I \) represents the maximum distance between $wp^c_i$ and  $wp^c_{i+1}$, while 
\(  d_r \) denotes the triggering distance at which the maximum value of \(   ||wp^c_{i+1}-wp^c_{i}|| \) is shrunk. 
In addition, the length of the sub-path $wp^c$ is defined as \(  d_{wp}^c \), with its maximum value \(  d_{l} \) (look-ahead distance), \emph{i.e.}, \(d_{wp}^c= \sum_{i=1} ^{n-1} ||wp^c_{i+1}-wp^c_{i}|| \).
The functions for above constraints can be briefly described as follows:
\begin{enumerate}
   \item Equation (\ref{eq33}) establishes the connection between waypoints in the sub-path by setting a maximum distance constraint between them. However, if the waypoints approach the vicinity of the goal, the maximum distance is reduced to avoid overshooting the goal point.
   \item  Equation (\ref{eq34}) defines the maximum look-ahead distance for each sub-path. 
   If a sub-path is approaching the goal, the maximum look-ahead distance \(d_{l}\) is also reduced accordingly with \(d_{I}\) to prevent overshooting the goal point. 
   This adjustment is made to ensure that the path planning algorithm stays within the desired proximity of the goal.
\end{enumerate}
\begin{figure}[!h]
	\hspace*{-0.0cm}
	\vspace*{-0.0cm}
	\includegraphics[width=0.5\textwidth]{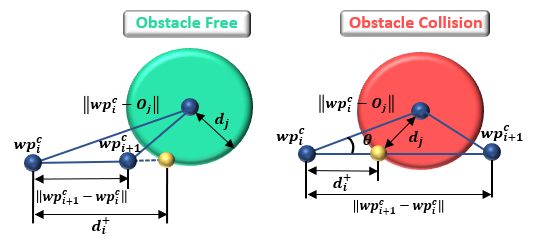}
	\caption{Illustration of the obstacle avoidance for sub-path}
	\label{fig:obs_avoid}
\end{figure}
\subsubsection{Obstacle Avoidance}
The search scope for waypoints is specifically confined within a circular region centered at \(wp^c_1\) and with a radius of \(d_l\)), focusing on the vicinity of \(wp^c_1\) rather than the obstacles across the entire map.
 To ensure that the generated sub-path does not cross any obstacles, we apply the following constraints:
{
{\begin{equation} \label{eq36}
||wp^c_{i}-O_j||  \geq d_j = r_j + d_{safe},
\end{equation}}}
{where {\(r_j\) calculates the radius of the \( j^{th}\) obstacle \( O_j\) with \(j \in \{1,2,\dots,n_{obs}\}\),}  inflation radius is defined by \(d_{safe}\) which determines the safe distance that should be maintained between obstacles and waypoints. 
Furthermore, to guarantee the generated path is obstacle-free, {where {the length between adjacent waypoints inside the sub-path must satisfy the following condition:}
{
\begin{equation} \label{eq37}
||wp^c_{i+1}-wp^c_{i}||  \leq d^+_i \;
\end{equation}}
{where \( d^+_i\) represents the threshold distance between \( wp^c_{i+1}\) and \( wp^c_{i}\) that crosses the obstacle, as illustrated in Fig.~\ref{fig:obs_avoid}.
Given the angle \(\theta\) between the obstacle \(O_j\) and waypoint \(wp_c^i\), \( d^+_i\) can be easily calculated by using the law of cosines:} }
{
\begin{equation*} \label{eq38}
d_j^2 = ||wp^c_{i}-O_j||^2  +  (d^+_i)^2 - 2||wp^c_{i}-O_j||\cdot  d^+_i \cdot \cos\theta
\end{equation*}}

\begin{algorithm}[!h]
    \SetKwInOut{Input}{Input}
    \SetKwInOut{Output}{Output}
    \Input{Start and goal locations: $start$, $goal$\;
    Fused memory layer of PPUM: $m^{'}$\;
    Geometry Map: Obstacles:$ $ $O$ $\in \mathbb{R}^{n_{obs} \times 1 }$ }
    {Tunable parameters: $d_I$,$d_l$,$d_r$,$d_s$,$n$}\;
    \Output{$Valid$ $Path$;}
    \textbf{Initialize}: $d_{e2g}^c= \infty$, $wp^c_{n}= start$, iteration number $c=1$ \;

\While{$d_{e2g}^c$ $>$ $d_s$}
      {
       
       Create series of random waypoints $wp^c$, $\in \mathbb{R}^{n \times 2 }$   \;
       Assigning probability cost to $(wp^c_i)$ :\\
           \For{each $wp^c_i$ $(i \in n)$ }{
               Probability of state “Crowded” is:
 
               $P(wp^c_{i})=GP(m^{'},wp^c_i)$    
 
               } 
       \textbf{minimize} solution from equation. (\ref{eq31})\; 
       \textbf{subject to} constraints from equation. {(\ref{eq33})-(\ref{eq37})}\;
       $d_{e2g}^c$ $\leftarrow$ $||wp_{n}^c-goal||$\;
          
       Add $wp_2^c$ to $Valid$ $Path$ \;
       $c=c+1$\;
      }
     
\caption{RHO for Path planning}

\label{alg:RH}
\end{algorithm}

\begin{figure}[!h]
\centering
	\hspace*{-0.0cm}
	\vspace*{-0.0cm}
	\includegraphics[width=0.45\textwidth]{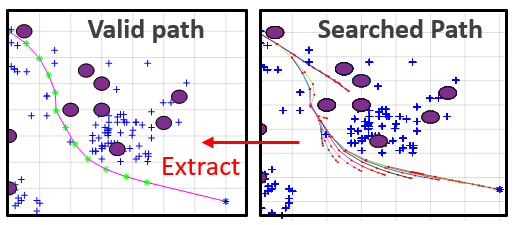}
	\caption{An example of RHO-based Path Planning, highlights the extraction of a valid path from the paths generated (pedestrians are represented by blue stars, while purple cylinders representing obstacles in the environment).}
	\label{fig:rhoplanning}
\end{figure}

\subsection{Path Planning Method}\label{subsec ppm}
The path planning problem, formulated with the introduced key formulations, can be approached as a  RHO problem. In this work, we employ CasADi \cite{Andersson2019} as our optimization framework, which is a versatile and powerful software framework that provides flexibility in modeling and solving optimization problems. It offers a wide range of capabilities and efficient solvers, making it suitable for tackling complex path planning problems. The path planning algorithm is explicitly presented in Algorithm \ref{alg:RH}. 
The “\(GP\)” function extracts the probability information from the PPUM. 
Specifically, the algorithm exhibits the following key features:
\begin{enumerate}
   \item The global planning problem is divided into sub-problems, which are solved sequentially. The optimization process starts from a fixed location, which is the second node of the previous path. The optimization continues until the end node  \(wp_n\) of the path reaches within the inflation radius \(d_s\) of the goal point. This condition ensures that the generated path is sufficiently close to the goal. When this condition is met, the optimization process stops, and the final path can be obtained (see Line 3 in Algorithm \ref{alg:RH}). 
    \item The valid path consists of all the waypoints from the second node of each path, which is obtained after the each iteration. By considering only the second nodes, the path ensures smoothness and continuity while adhering to the constraints and objectives of the path planning algorithm (see Line 13 in Algorithm \ref{alg:RH}).
    \item The performance of path planning is primarily influenced by two key parameters: the number of waypoints (\( n \)) and the lookahead distance (\( d_l \)). Increasing the number of waypoints provides greater flexibility in shaping the path but can lead to slower computational speed. 
    On the other hand, increasing the lookahead distance allows the generated path to be more adaptive to complex scenarios by extending the horizon scope. Note that when increasing \( d_l \), \( d_I \) should also be increased without changing \( n \). A suggested setting for \( d_I \) could follow \( d_I = d_l/n \).
\end{enumerate}

Finally, an example of the proposed path planning method and the illustration of path generation can be seen in Fig.~\ref{fig:rhoplanning} which showcases how the planned path (purple curve in the left figure) is generated in a receding horizon manner, adapting to a complex environment.

\section{Simulation Studies}\label{secsim}
In order to analyze and evaluate the effectiveness of the proposed RHO path planning with PPUM, the method's performance is evaluated in the following cases:
\begin{itemize}
    \item  Case 1: Comparative study between the proposed probability-based memory fusion method (PPUM), a time-dependent memory fusion strategy (PUM) described in the previous study \cite{gecongestion} and  an off-line memory model (OLM) without updates, focusing on the accuracy of crowd spatial-temporal estimation.
    \item  Case 2: Comparative study of the proposed RHO-based path planning strategy with two benchmark congestion-aware routing methods introduced in \cite{gecongestion} and \cite{Zhang2020a}, as well as a classic A* algorithm.  All the planning algorithms utilized the same memory information in this case. The study mainly focuses on evaluating their performance in terms of congestion avoidance and path efficiency.
   \item  Case 3: Comparative study of the overall performance between the proposed path planning framework (RHO-based path planning with PPUM), our previous framework (Congestion-aware A* with PUM) presented in \cite{gecongestion}, and a non-update framework (Congestion-aware A* with OLM). The study aims to evaluate their overall performance in terms of congestion awareness, and path optimality.  
\end{itemize}

\subsection{Case 1: Memory fusion performance evaluation}
\begin{figure}[hbt!]
    \centering
	\hspace*{-0.0cm}
	\vspace*{-0.0cm}
	\includegraphics[width=0.3\textwidth]{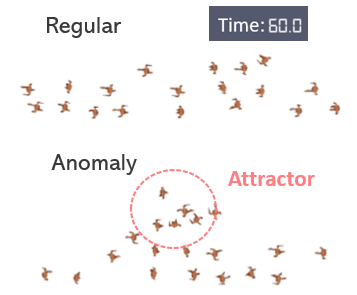}
	\caption{{A comparison of two virtual environments with the same initial settings at 60 seconds after simulation, where one environment contains an anomaly.}}
	\label{fig:sim_anomoly}
\end{figure}

\subsubsection{Setup} The crowd's macroscopic spatial-temporal behaviour is simulated in Pathfinder, a crowd simulator introduced in \cite{thornton2011pathfinder}.  
This simulator provides a platform to generate realistic crowd movements and interactions, allowing for the evaluation and analysis of the proposed memory fusion algorithms in various crowd scenarios. 

{In the simple scenario of a corridor area with bidirectional crowd flow, depicted in Fig.~\ref{fig:sim_anomoly}, initially the crowds were simulated to pass through the corridor every 10 seconds. The passing speed and crowd size are randomly generated based on normal distributions, representing the regular behavior of the crowd in the corridor.}
In addition, an invisible circular attractor is placed at the center of the corridor. 
When pedestrians enter the influence radius of the attractor, they are attracted towards it and briefly stay nearby before continuing their original route. 
This attractor introduces an anomaly event in the simulation environment, disrupting the original regularity of the crowd's behavior. 
The presence of this anomaly is illustrated in Fig.~\ref{fig:sim_anomoly}. Initially, both the PUM and the PPUM are initialized with identical regular patterns. Subsequently, an observed anomaly is introduced into both memory models by considering those extra coordinates that deviate from the regular patterns. For simplicity, we extract the coordinates of anomalous pedestrians to represent sensor tracking results in the PUM model. In the case of the PPUM model, which requires a GMM input, we manually introduce some white noise. This experimental setup allows us to investigate how the accuracy of the two memory models' performance evolves over time in response to the introduced anomaly.

\subsubsection{Memory fusion results}In the presence of an anomaly, the evaluation is conducted by calculating the average Root Mean Square Error (RMSE) between the mean values obtained from different memory sources and the ground truth crowd distribution. 
The performance of each memory source over 80 seconds after observation of the anomaly can be seen in Fig.~\ref{fig:sim_c1}. 
Generally, the results presented in TABLE \ref{tab:1} indicate that the proposed PPUM method generates more accurate crowd spatial-temporal estimations, particularly when unexpected events occur. 
However, it is worth noting that in Anomaly 2, where the anomalous event ends before the simulation duration, the PUM method shows better adaptability due to its time-dependent nature. 
This implies that the estimation accuracy of PPUM needs to be guaranteed through a high update frequency. 
Even when the anomaly disappears, the overall accuracy of PPUM still outperforms the other two memory sources, indicating its effectiveness in capturing and representing crowd behavior.

\begin{table}[!h]
\caption{Case 1:Average RMSE of different memory models in terms of crowd distribution reasoning under anomaly. }
\begin{adjustbox}{width=2.5in,center}
\centering
\begin{tabular}{|c|c|c|}
\hline
Average RMSE & Anomaly 1 & Anomaly 2 \\ \hline
OLM          & 0.0118    & 0.0112    \\ \hline
PUM          & 0.0098    & 0.0086    \\ \hline
PPUM         & 0.0076    & 0.0083    \\ \hline
\end{tabular}
\end{adjustbox}
\label{tab:1}
\end{table}

\begin{figure}[H]
    \centering
	\hspace*{-0.0cm}
	\vspace*{-0.0cm}
	\includegraphics[width=0.4\textwidth]{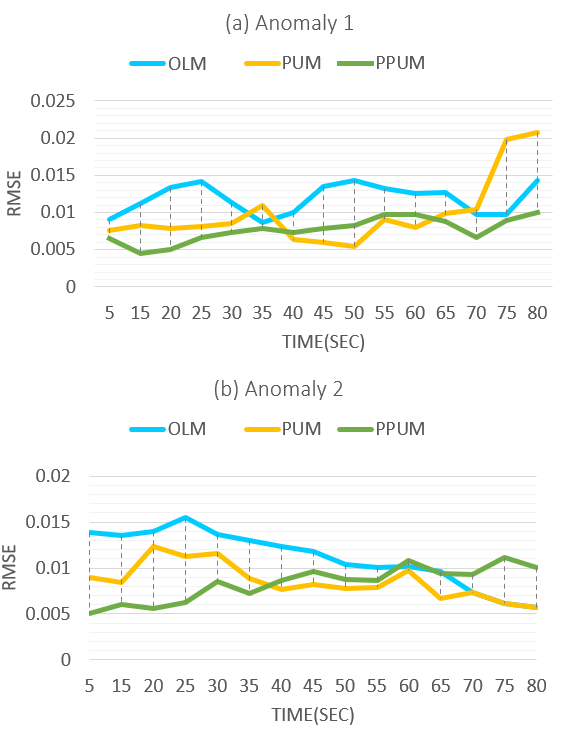}
	\caption{Case 1: Estimation errors of OLM, PUM, and PPUM: (a) Anomaly 1: The attractor exists throughout the entire simulation. The probability-based reasoning feature of PPUM enables its overall approximation to be closer to the ground truth data. (b) Anomaly 2: The attractor is removed at 30 seconds into the simulation. As the observed anomaly fades in PUM with time, the estimation gradually converges to the original OLM (visible in the overlaying line during 70-80 seconds). In this case, the average error of PPUM is only slightly lower than that of the other two memory models.}
	\label{fig:sim_c1}
\end{figure}

\subsection{Case 2: Path Planning Performance}

\subsubsection{Setup} The comparative study of different path planning methods is conducted in a \(20m \times 20m \) space, where random obstacles and crowds are generated. The start point and goal point remain fixed for each test scenario. The purpose of this study is to evaluate the performance of different individual planning algorithms under varying crowd sizes.

\subsubsection{Evaluation method}Congestion-aware path planning methods commonly incorporate detours to bypass congested areas. 
However, determining whether these detours actually lead to time savings or delays presents a significant challenge in real-world scenarios before the travel is finished. 
The outcome of this evaluation depends on several crucial factors, including the size of the robot and the capabilities of its local motion planning algorithm.

To assess the performance of path planning in such cases, a travel time estimator is employed for the global reference path. 
This estimator utilizes a crowd spatial distribution map to approximate the expected travel time for the calculated path. 
The process is illustrated in Fig.~\ref{fig:exptime}. 
Initially, a virtual corridor with a width of one meter is created along the path. 
The crowd size within this virtual corridor is measured, and subsequently, the travel time is approximated using the travel time model described in \cite{trautman2015robot}. 
This model employs multi-goal interacting Gaussian processes as the local planner. 

\begin{figure}[ht]
    \centering
	\hspace*{-0.0cm}
	\vspace*{-0.0cm}
	\includegraphics[width=0.25\textwidth]{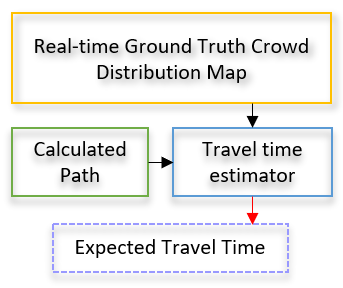}
	\caption{Flow chart of how to calculate the expected travel time }
	\label{fig:exptime}
\end{figure}

To quantify the improvement of RHO-based path planning method compared to other benchmark methods in terms of path efficiency, an evaluation index is defined as follows:
\begin{equation*} \label{eq38}
T_s = {\frac{T_{ben}-T_{RHO}}{T_{ben}}}
\end{equation*}
where \(T_{s}\) represents improvement of the path efficiency, calculated by the time saved by comparing the travel time from our method \(T_{RHO}\) with the travel time from the benchmark method \(T_{ben}\).

\subsubsection{Path planning result}
During the conducted 30 tests, we evaluated the improvement of the proposed method compared to other benchmark methods for different crowd sizes. The results, as explicitly shown in Table \ref{tab:2}, demonstrate that as the crowd size increases, the improvement achieved by the proposed method becomes more significant. This indicates that the effectiveness of the proposed method in enhancing path efficiency is more pronounced in scenarios with larger crowds. An example of the result can be viewed in Fig. \ref{fig:case2}.
\begin{figure}[!h]
    \centering
	\hspace*{-0.0cm}
	\vspace*{-0.0cm}
	\includegraphics[width=0.35\textwidth]{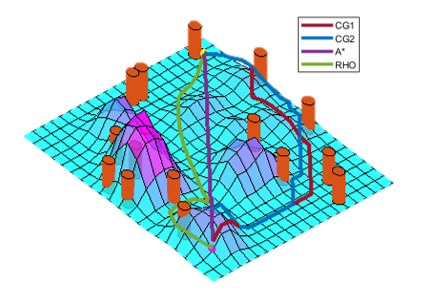}
	\caption{An example of Case 2 illustrating the performance of different algorithms in the presence of crowds and obstacles (represented by orange cylinders).}
	\label{fig:case2}
\end{figure}

\begin{table*}[]

\caption{Case 2: Simulation results depicting the average estimated travel time for different path planning algorithms across 30 tests, considering varying crowd sizes. CG1 and CG2 are the congestion-aware path planning algorithm introduced in \cite{Zhang2020a} and \cite{gecongestion} respectively, RHO represents the proposed RHO for path planning.}
\centering
\begin{tabular}{|ccccccccccccc|}
\hline
\multicolumn{13}{|c|}{Estimated Travel Time to Destination (sec)}                                                                                                                                                                                                                                                                                                                                                                                       \\ \hline
\multicolumn{1}{|c|}{\multirow{2}{*}{\begin{tabular}[c]{@{}c@{}}Map \\ setting\end{tabular}}} & \multicolumn{12}{c|}{obstacle number: 15}                                                                                                                                                                                                                                                                                                               \\ \cline{2-13} 
\multicolumn{1}{|c|}{}                                                                        & \multicolumn{4}{c|}{crowd size: 30}                                                                                     & \multicolumn{4}{c|}{crowd size: 60}                                                                                     & \multicolumn{4}{c|}{crowd size: 100}                                                                \\ \hline
\multicolumn{1}{|c|}{Method}                                                                  & \multicolumn{1}{c|}{CG1}     & \multicolumn{1}{c|}{CG2}     & \multicolumn{1}{c|}{A*}      & \multicolumn{1}{c|}{RHO}   & \multicolumn{1}{c|}{CG1}     & \multicolumn{1}{c|}{CG2}     & \multicolumn{1}{c|}{A*}      & \multicolumn{1}{c|}{RHO}   & \multicolumn{1}{c|}{CG1}     & \multicolumn{1}{c|}{CG2}     & \multicolumn{1}{c|}{A*}      & RHO    \\ \hline
\multicolumn{1}{|c|}{Average}                                                                 & \multicolumn{1}{c|}{106.23}  & \multicolumn{1}{c|}{106.31}  & \multicolumn{1}{c|}{118.52}  & \multicolumn{1}{c|}{88.28} & \multicolumn{1}{c|}{141.24}  & \multicolumn{1}{c|}{131.5}   & \multicolumn{1}{c|}{153.63}  & \multicolumn{1}{c|}{97.23} & \multicolumn{1}{c|}{179.51}  & \multicolumn{1}{c|}{163.09}  & \multicolumn{1}{c|}{212.22}  & 111.59 \\ \hline
\multicolumn{1}{|c|}{\(T_s\)}                                                                       & \multicolumn{1}{c|}{16.90\%} & \multicolumn{1}{c|}{16.95\%} & \multicolumn{1}{c|}{25.51\%} & \multicolumn{1}{c|}{}      & \multicolumn{1}{c|}{31.16\%} & \multicolumn{1}{c|}{26.06\%} & \multicolumn{1}{c|}{36.71\%} & \multicolumn{1}{c|}{}      & \multicolumn{1}{c|}{37.83\%} & \multicolumn{1}{c|}{31.57\%} & \multicolumn{1}{c|}{47.41\%} &        \\ \hline
\end{tabular}
\label{tab:2}
\end{table*}

\subsection{Case 3: Overall Performance}
\begin{figure}[h]
    \centering
	\hspace*{-0.0cm}
	\vspace*{-0.0cm}
	\includegraphics[width=0.25\textwidth]{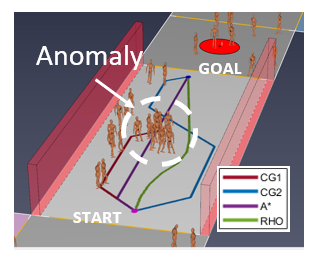}
	\caption{An example of Case 3.1 is presented, showcasing a corridor with bidirectional pedestrian flow. The figure illustrates the generated paths from different frameworks. The paths are overlaid on the corridor map, providing a visual comparison of the planned routes in the presence of bidirectional pedestrian flow with anomaly.}
	\label{fig:case3111}
\end{figure}
\subsubsection{Case 3.1}
In this case, a small-scale scenario is designed to evaluate the performance of memory-integrated path planning algorithms. 
The scenario consists of a corridor with bidirectional crowd flow and the presence of one observed anomaly, following the same setting described in Case 1. 
By using this controlled environment, we can effectively assess the capabilities of different methods in planning under constrained spaces and responding to unexpected events. The start and goal points are randomly generated from the edges of the corridor. 
Different methods are then employed to calculate a path every 1 minute during a 20-minute simulation. 
An example of the resulting path can be seen in Fig.~\ref{fig:case3111}. 
The simulation results are presented in Table \ref{tab:3}. 
It can be observed that before the anomaly ended, there was a significant improvement compared to other benchmark methods. However, as there was no anomaly during the 10-20 minute period, the crowd spatial distribution gradually recovered to its original pattern. As a result, Method 2 caught up to our method due to its time-dependent changing feature. On average, our method showed a \(25.24\%\) improvement compared to our previous framework throughout the entire simulation. Furthermore, our method largely outperformed the other two frameworks in scenarios with observed anomalies.

\subsubsection{Case 3.2}
\begin{figure}[H]
    \centering
	\hspace*{-0.0cm}
	\vspace*{-0.0cm}
	\includegraphics[width=0.4\textwidth]{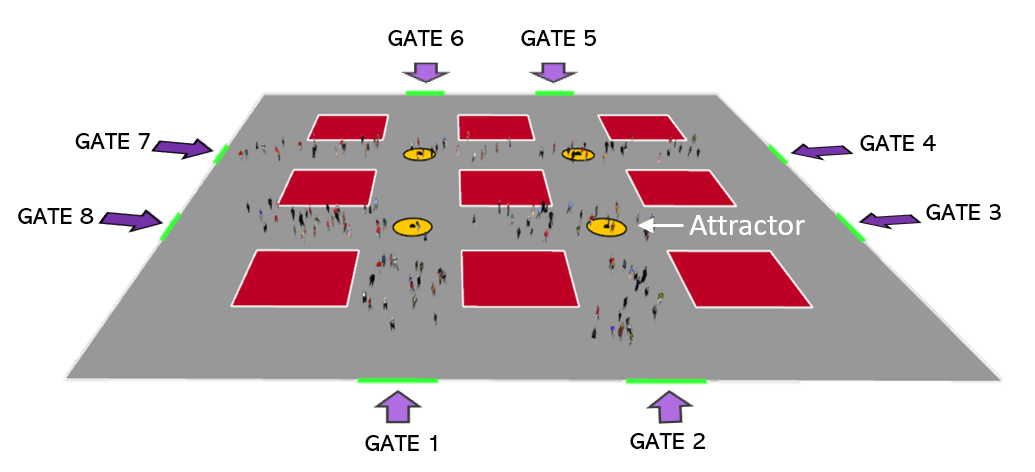}
	\caption{Environment configuration of Case 3.2: Crowd 1 enters from GATE 1 and exits at GATE 6. Crowd 2 enters from GATE 5 and exits at GATE 2. Crowd 3 enters from GATE 5 and exits at GATE 3. Crowd 4 enters from GATE 4 and exits at GATE 7. {All crowds were set to enter the map via these gates every 3 minutes,}  simulating the crowd's spatial-temporal regularity. Four attractors are symmetrically placed at the cross points of the map, indicated by yellow circles. The red cubes represent the walls.}
	\label{fig:case32env}
\end{figure}
In this case, we established a \(70m \times 70m\) map, as shown in Fig.~\ref{fig:case32env}. 
The map contains four attractors strategically placed to disturb the normal crowd behavior, representing anomalies in the uncertain environment. 
These attractors are randomly activated during the 60-minute simulations and are assumed to be observed by the sensors. 
The purpose of this setup is to create a challenging environment where the the calculated global path reference need to effectively respond to these observed anomalies. 
As indicated in Table \ref{tab:4}, even without observation updates, the proposed framework still demonstrates a slight improvement in the individual performance of the RHO planning algorithm compared to other benchmark methods. 
In additional, it is evidented that as the number of observed anomalies increases, the proposed RHO+PPUM method consistently outperforms other benchmark methods, demonstrating its high adaptability to uncertain environments with sufficient observation updates. 
One simulation result is illustrated in Fig.~\ref{fig:case23r}. 
\begin{table}[]
\caption{Simulation results of Case 3.1 are presented, focusing on the path planning performance in a small-scale area at different periods of one observed anomaly. The anomaly is removed at 10 minutes, but it still has a short-term “tail effect” on the crowd before they return to their original routes.}
\centering
\begin{tabular}{|ccccc}
\hline
\multicolumn{5}{|c|}{Estimated Travel Time to Destination (sec)}                                                                                        \\ \hline
\multicolumn{1}{|c|}{Time (min)}    & \multicolumn{1}{c|}{CG1}     & \multicolumn{1}{c|}{CG2}     & \multicolumn{1}{c|}{A*}      & \multicolumn{1}{c|}{RHO}   \\ \hline
\multicolumn{1}{|c|}{5}       & \multicolumn{1}{c|}{192.1}   & \multicolumn{1}{c|}{120.3}   & \multicolumn{1}{c|}{278.78}  & \multicolumn{1}{c|}{78.24} \\ \hline
\multicolumn{1}{|c|}{10}      & \multicolumn{1}{c|}{181.63}  & \multicolumn{1}{c|}{131.63}  & \multicolumn{1}{c|}{272.38}  & \multicolumn{1}{c|}{77.26} \\ \hline
\multicolumn{1}{|c|}{15}      & \multicolumn{1}{c|}{98.29}   & \multicolumn{1}{c|}{70.69}   & \multicolumn{1}{c|}{176.29}  & \multicolumn{1}{c|}{60.03} \\ \hline
\multicolumn{1}{|c|}{20}      & \multicolumn{1}{c|}{94.54}   & \multicolumn{1}{c|}{73.77}   & \multicolumn{1}{c|}{165.95}  & \multicolumn{1}{c|}{64.62} \\ \hline
\multicolumn{1}{|c|}{Time (min)}    & \multicolumn{3}{c|}{Efficiency improvement $T_s$}                                           &                            \\ \cline{1-4}
\multicolumn{1}{|c|}{5}       & \multicolumn{1}{c|}{59.27\%} & \multicolumn{1}{c|}{34.92\%} & \multicolumn{1}{c|}{71.19\%} &                            \\ \cline{1-4}
\multicolumn{1}{|c|}{10}      & \multicolumn{1}{c|}{57.46\%} & \multicolumn{1}{c|}{38.54\%} & \multicolumn{1}{c|}{74.21\%} &                            \\ \cline{1-4}
\multicolumn{1}{|c|}{15}      & \multicolumn{1}{c|}{38.75\%} & \multicolumn{1}{c|}{15.08\%} & \multicolumn{1}{c|}{65.95\%} &                            \\ \cline{1-4}
\multicolumn{1}{|c|}{20}      & \multicolumn{1}{c|}{31.65\%} & \multicolumn{1}{c|}{12.40\%} & \multicolumn{1}{c|}{61.06\%} &                            \\ \cline{1-4}
\multicolumn{1}{|c|}{Average} & \multicolumn{1}{c|}{46.78\%} & \multicolumn{1}{c|}{25.24\%} & \multicolumn{1}{c|}{68.10\%} &                            \\ \cline{1-4}
\end{tabular}

\label{tab:3}
\end{table}

\begin{table}[]
\caption{Simulation results of Case 3.2 are presented, focusing on the path efficiency improvement. \(N_{a}\)  represents the number of observed anomalies. The average \(T_s\) is the mean value calculated from three path samples, indicating the overall improvement in estimated travel time compared to the benchmark methods.}
\centering
\begin{tabular}{|c|ccccc|}
\hline
Average $T_s$ & \multicolumn{5}{c|}{Proposed   RHO+PPUM frame work}                                                                                 \\ \hline
$N_{a}$         & \multicolumn{1}{c|}{0}       & \multicolumn{1}{c|}{5}       & \multicolumn{1}{c|}{10}      & \multicolumn{1}{c|}{15}      & 20      \\ \hline
CG1+OLM        & \multicolumn{1}{c|}{12.67\%} & \multicolumn{1}{c|}{29.45\%} & \multicolumn{1}{c|}{36.90\%} & \multicolumn{1}{c|}{43.51\%} & 48.77\% \\ \hline
CG2+PUM        & \multicolumn{1}{c|}{11.23\%} & \multicolumn{1}{c|}{15.73\%} & \multicolumn{1}{c|}{23.41\%} & \multicolumn{1}{c|}{26.65\%} & 37.66\% \\ \hline
A*             & \multicolumn{1}{c|}{24.90\%} & \multicolumn{1}{c|}{33.23\%} & \multicolumn{1}{c|}{44.21\%} & \multicolumn{1}{c|}{49.39\%} & 52.23\% \\ \hline
\end{tabular}

\label{tab:4}
\end{table}

\begin{figure}[H]
    \centering
	\hspace*{-0.0cm}
	\vspace*{-0.0cm}
	\includegraphics[width=0.3\textwidth]{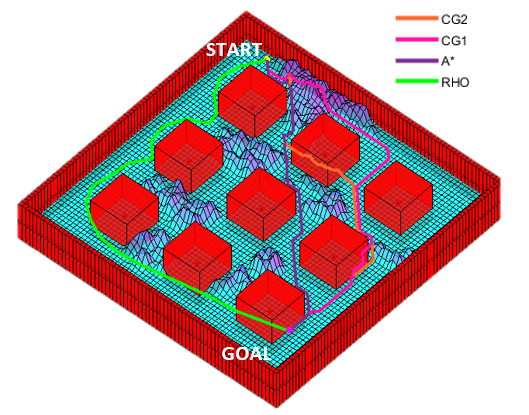}
	\caption{An example of the simulation results for Case 3.2:  different path performances under the ground truth crowd density map are displayed.}
	\label{fig:case23r}
\end{figure}

\section{Conclusion}\label{seccon}
The presented study introduces a probability-based memory model for reasoning the spatial distribution of crowds in uncertain environments. The model incorporates prior knowledge from prediction and likelihood evidence from sensor observations, combining them using a novel weighted Evidence D-S rule. Additionally, a RHO-based path planning method is proposed to integrate the memory model and provide optimal path references for robots. The comparative simulations demonstrate that the PPUM estimation closely approximates the real crowd spatial distribution, while the proposed path planning method exhibits higher efficiency and adaptability to congestion areas. The findings in Case 3 validate the superiority and feasibility of the PPUM+RHO framework over other benchmark frameworks, showcasing improved path efficiency, adaptability to uncertain environments, and congestion avoidance.

The probability-based reasoning method is identified as a key component for adaptive path planning in uncertain environments. However, the current memory fusion method only considers the sensor observation area, and future work will concentrate on reasoning the unobserved areas potentially affected by anomalies based on limited sensor observations. Additionally, real-world experiments are necessary to validate the performance of the presented work.

\ifCLASSOPTIONcaptionsoff
  \newpage
\fi



\bibliographystyle{IEEEtran}
\bibliography{mybib}

\begin{thebibliography}{10}
\providecommand{\url}[1]{#1}
\csname url@samestyle\endcsname
\providecommand{\newblock}{\relax}
\providecommand{\bibinfo}[2]{#2}
\providecommand{\BIBentrySTDinterwordspacing}{\spaceskip=0pt\relax}
\providecommand{\BIBentryALTinterwordstretchfactor}{4}
\providecommand{\BIBentryALTinterwordspacing}{\spaceskip=\fontdimen2\font plus
\BIBentryALTinterwordstretchfactor\fontdimen3\font minus
  \fontdimen4\font\relax}
\providecommand{\BIBforeignlanguage}[2]{{%
\expandafter\ifx\csname l@#1\endcsname\relax
\typeout{** WARNING: IEEEtran.bst: No hyphenation pattern has been}%
\typeout{** loaded for the language `#1'. Using the pattern for}%
\typeout{** the default language instead.}%
\else
\language=\csname l@#1\endcsname
\fi
#2}}
\providecommand{\BIBdecl}{\relax}
\BIBdecl

\bibitem{lin2020spatiotemporal}
C.~Lin, G.~Han, J.~Du, T.~Xu, L.~Shu, and Z.~Lv, ``Spatiotemporal
  congestion-aware path planning toward intelligent transportation systems in
  software-defined smart city iot,'' \emph{IEEE Internet of Things Journal},
  vol.~7, no.~9, pp. 8012--8024, 2020.

\bibitem{morales2013human}
Y.~Morales, N.~Kallakuri, K.~Shinozawa, T.~Miyashita, and N.~Hagita,
  ``Human-comfortable navigation for an autonomous robotic wheelchair,'' in
  \emph{IEEE/RSJ International Conference on Intelligent Robots and Systems},
  2013, pp. 2737--2743.

\bibitem{tokekar2014energy}
P.~Tokekar, N.~Karnad, and V.~Isler, ``Energy-optimal trajectory planning for
  car-like robots,'' \emph{Autonomous Robots}, vol.~37, pp. 279--300, 2014.

\bibitem{Krajnik2014}
T.~Krajnik, J.~P. Fentanes, G.~Cielniak, C.~Dondrup, and T.~Duckett, ``Spectral
  analysis for long-term robotic mapping,'' in \emph{IEEE International
  Conference on Robotics and Automation}, 2014, pp. 3706--3711.

\bibitem{Fentanes2017}
T.~Krajn{\'\i}k, J.~P. Fentanes, J.~M. Santos, and T.~Duckett, ``Fremen:
  Frequency map enhancement for long-term mobile robot autonomy in changing
  environments,'' \emph{IEEE Transactions on Robotics}, vol.~33, no.~4, pp.
  964--977, 2017.

\bibitem{Kiss2021}
S.~H. Kiss, K.~Katuwandeniya, A.~Alempijevic, and T.~Vidal-Calleja,
  ``Probabilistic dynamic crowd prediction for social navigation,'' in
  \emph{IEEE International Conference on Robotics and Automation}, 2021, pp.
  9269--9275.

\bibitem{miglani2019deep}
A.~Miglani and N.~Kumar, ``Deep learning models for traffic flow prediction in
  autonomous vehicles: A review, solutions, and challenges,'' \emph{Vehicular
  Communications}, vol.~20, p. 100184, 2019.

\bibitem{Luo2019}
X.~Luo, D.~Li, Y.~Yang, and S.~Zhang, ``Spatiotemporal traffic flow prediction
  with knn and lstm,'' \emph{Journal of Advanced Transportation}, vol. 2019,
  2019.

\bibitem{Chen2020a}
C.~Chen, J.~Jiang, N.~Lv, and S.~Li, ``An intelligent path planning scheme of
  autonomous vehicles platoon using deep reinforcement learning on network
  edge,'' \emph{IEEE Access}, vol.~8, pp. 99\,059--99\,069, 2020.

\bibitem{hojati2013hazard}
A.~T. Hojati, L.~Ferreira, S.~Washington, and P.~Charles, ``Hazard based models
  for freeway traffic incident duration,'' \emph{Accident Analysis \&
  Prevention}, vol.~52, pp. 171--181, 2013.

\bibitem{nam2000exploratory}
D.~Nam and F.~Mannering, ``An exploratory hazard-based analysis of highway
  incident duration,'' \emph{Transportation Research Part A: Policy and
  Practice}, vol.~34, no.~2, pp. 85--102, 2000.

\bibitem{junhua2013estimating}
W.~Junhua, C.~Haozhe, and Q.~Shi, ``Estimating freeway incident duration using
  accelerated failure time modeling,'' \emph{Safety science}, vol.~54, pp.
  43--50, 2013.

\bibitem{gecongestion}
Z.~Ge, J.~Jiang, and M.~Coombes, ``A congestion-aware path planning method
  considering crowd spatial-temporal anomalies for long-term autonomy of mobile
  robots,'' in \emph{IEEE International Conference on Robotics and Automation},
  2023, pp. 7930--7936.

\bibitem{chen2019crowd}
C.~Chen, Y.~Liu, S.~Kreiss, and A.~Alahi, ``Crowd-robot interaction:
  Crowd-aware robot navigation with attention-based deep reinforcement
  learning,'' in \emph{IEEE International Conference on Robotics and
  Automation}, 2019, pp. 6015--6022.

\bibitem{nishimura2020l2b}
M.~Nishimura and R.~Yonetani, ``L2b: Learning to balance the safety-efficiency
  trade-off in interactive crowd-aware robot navigation,'' in \emph{IEEE/RSJ
  International Conference on Intelligent Robots and Systems}.\hskip 1em plus
  0.5em minus 0.4em\relax IEEE, 2020, pp. 11\,004--11\,010.

\bibitem{gao2022evaluation}
Y.~Gao and C.-M. Huang, ``Evaluation of socially-aware robot navigation,''
  \emph{Frontiers in Robotics and AI}, vol.~8, p. 420, 2022.

\bibitem{Vega2019}
A.~Vega, L.~J. Manso, D.~G. Macharet, P.~Bustos, and P.~N{\'u}{\~n}ez,
  ``Socially aware robot navigation system in human-populated and interactive
  environments based on an adaptive spatial density function and space
  affordances,'' \emph{Pattern Recognition Letters}, vol. 118, pp. 72--84,
  2019.

\bibitem{Zhang2020a}
Z.~Zhang, H.~Liu, Z.~Jiao, Y.~Zhu, and S.-C. Zhu, ``Congestion-aware evacuation
  routing using augmented reality devices,'' in \emph{IEEE International
  Conference on Robotics and Automation}, 2020, pp. 2798--2804.

\bibitem{basu2003routing}
A.~Basu, A.~Lin, and S.~Ramanathan, ``Routing using potentials: a dynamic
  traffic-aware routing algorithm,'' in \emph{Proceedings of the conference on
  Applications, technologies, architectures, and protocols for computer
  communications}, 2003, pp. 37--48.

\bibitem{dodd2005role}
W.~Dodd and R.~Gutierrez, ``The role of episodic memory and emotion in a
  cognitive robot,'' in \emph{IEEE International Workshop on Robot and Human
  Interactive Communication.}, 2005, pp. 692--697.

\bibitem{Zou2016}
Q.~Zou, D.~Liu, M.~Cong, Y.~Cui, and Y.~Du, ``Robotic cognitive map building
  based on biology-inspired memory,'' in \emph{IEEE International Conference on
  Robotics and Biomimetics}, 2016, pp. 1814--1819.

\bibitem{persiani2018working}
M.~Persiani, A.~M. Franchi, and G.~Gini, ``A working memory model improves
  cognitive control in agents and robots,'' \emph{Cognitive Systems Research},
  vol.~51, pp. 1--13, 2018.

\bibitem{welch1995introduction}
G.~Welch and G.~Bishop, ``An introduction to the kalman filter,'' 1995.

\bibitem{Fentanes2015}
J.~P. Fentanes, B.~Lacerda, T.~Krajn{\'\i}k, N.~Hawes, and M.~Hanheide, ``Now
  or later? predicting and maximising success of navigation actions from
  long-term experience,'' in \emph{IEEE International Conference on Robotics
  and Automation}, 2015, pp. 1112--1117.

\bibitem{B2018}
S.~Molina, G.~Cielniak, T.~Krajn{\'\i}k, and T.~Duckett, ``Modelling and
  predicting rhythmic flow patterns in dynamic environments,'' in \emph{Towards
  Autonomous Robotic Systems: 19th Annual Conference}.\hskip 1em plus 0.5em
  minus 0.4em\relax Springer, 2018, pp. 135--146.

\bibitem{Vintr2019}
T.~Vintr, Z.~Yan, T.~Duckett, and T.~Krajn{\'\i}k, ``Spatio-temporal
  representation for long-term anticipation of human presence in service
  robotics,'' in \emph{IEEE International Conference on Robotics and
  Automation}, 2019, pp. 2620--2626.

\bibitem{Krajnik2019}
T.~Krajn{\'\i}k, T.~Vintr, S.~Molina, J.~P. Fentanes, G.~Cielniak, O.~M. Mozos,
  G.~Broughton, and T.~Duckett, ``Warped hypertime representations for
  long-term autonomy of mobile robots,'' \emph{IEEE Robotics and Automation
  Letters}, vol.~4, no.~4, pp. 3310--3317, 2019.

\bibitem{dempster2008upper}
A.~P. Dempster \emph{et~al.}, ``Upper and lower probabilities induced by a
  multivalued mapping.'' \emph{Classic works of the Dempster-Shafer theory of
  belief functions}, vol. 219, no.~2, pp. 57--72, 2008.

\bibitem{dempster1968generalization}
A.~P. Dempster, ``A generalization of bayesian inference,'' \emph{Journal of
  the Royal Statistical Society: Series B (Methodological)}, vol.~30, no.~2,
  pp. 205--232, 1968.

\bibitem{shafer1976mathematical}
G.~Shafer, \emph{A mathematical theory of evidence}.\hskip 1em plus 0.5em minus
  0.4em\relax Princeton university press, 1976, vol.~42.

\bibitem{Li2014}
L.~Li, D.~Zhu, B.~Sun, and Z.~Deng, ``The 3-d map building of auv based on ds
  information fusion,'' in \emph{Proceedings of the 33rd Chinese Control
  Conference}.\hskip 1em plus 0.5em minus 0.4em\relax IEEE, 2014, pp.
  8639--8644.

\bibitem{Chen2018}
B.-C. Chen, X.~Tao, M.-R. Yang, C.~Yu, W.-M. Pan, and V.~C. Leung, ``A saliency
  map fusion method based on weighted ds evidence theory,'' \emph{IEEE Access},
  vol.~6, pp. 27\,346--27\,355, 2018.

\bibitem{Zhu2014}
D.~Zhu, W.~Li, M.~Yan, and S.~X. Yang, ``The path planning of auv based on ds
  information fusion map building and bio-inspired neural network in unknown
  dynamic environment,'' \emph{International Journal of Advanced Robotic
  Systems}, vol.~11, no.~3, p.~34, 2014.

\bibitem{Chen2019}
L.~Chen, L.~Diao, and J.~Sang, ``A novel weighted evidence combination rule
  based on improved entropy function with a diagnosis application,''
  \emph{International Journal of Distributed Sensor Networks}, vol.~15, no.~1,
  2019.

\bibitem{Beynon2000}
M.~Beynon, B.~Curry, and P.~Morgan, ``{The Dempster-Shafer theory of evidence:
  An alternative approach to multicriteria decision modelling},'' \emph{Omega},
  vol.~28, no.~1, pp. 37--50, 2000.

\bibitem{yongsheng2002study}
Z.~Yongsheng, W.~Chengdong, and Z.~Youyun, ``Study of neural network based on
  method of weighted balance of evidence and its application to fault
  diagnosis,'' \emph{Journal of Mechanical Engineering}, vol.~38, no.~6, pp.
  66--71, 2002.

\bibitem{guo2006weighted}
H.~Guo and L.~Zhang, ``A weighted balance evidence theory for structural
  multiple damage localization,'' \emph{Computer methods in applied mechanics
  and engineering}, vol. 195, no. 44-47, pp. 6225--6238, 2006.

\bibitem{Andersson2019}
J.~A.~E. Andersson, J.~Gillis, G.~Horn, J.~B. Rawlings, and M.~Diehl,
  ``{CasADi} -- {A} software framework for nonlinear optimization and optimal
  control,'' \emph{Mathematical Programming Computation}, vol.~11, no.~1, pp.
  1--36, 2019.

\bibitem{thornton2011pathfinder}
C.~Thornton, R.~O’Konski, B.~Hardeman, and D.~Swenson, ``Pathfinder: an
  agent-based egress simulator,'' in \emph{Pedestrian and evacuation
  dynamics}.\hskip 1em plus 0.5em minus 0.4em\relax Springer, 2011, pp.
  889--892.

\bibitem{trautman2015robot}
P.~Trautman, J.~Ma, R.~M. Murray, and A.~Krause, ``Robot navigation in dense
  human crowds: Statistical models and experimental studies of human--robot
  cooperation,'' \emph{The International Journal of Robotics Research},
  vol.~34, no.~3, pp. 335--356, 2015.

\end{thebibliography}
%

%

\begin{IEEEbiography}[{\includegraphics[width=1in,height=1.25in,clip,keepaspectratio]{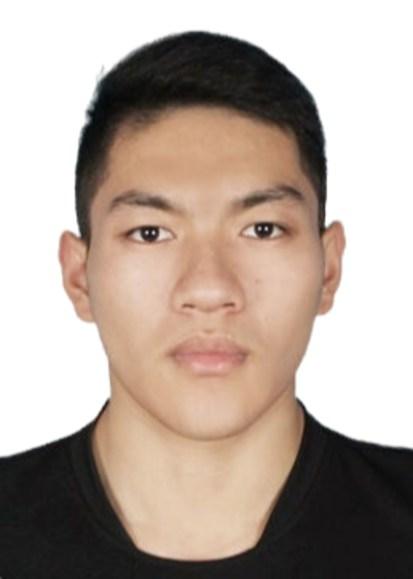}}]{Zijian Ge}
received the B.E.
degrees in automotive engineering from Tongji Zhejiang College, Jiaxing, China in 2020, and M.Sc. degree in automotive engineering from the Department of Aeronautical and Automotive Engineering, Loughborough University Leicester, U.K., in 2021, where he is  currently working toward the Ph.D degree at the Loughborough University Centre for Autonomous Systems. His research interests include autonomous vehicle path planning, human-robot interaction. 
\end{IEEEbiography}

\vskip 0pt plus -1fil

\begin{IEEEbiography}[{\includegraphics[width=1in,height=1.25in,clip,keepaspectratio]{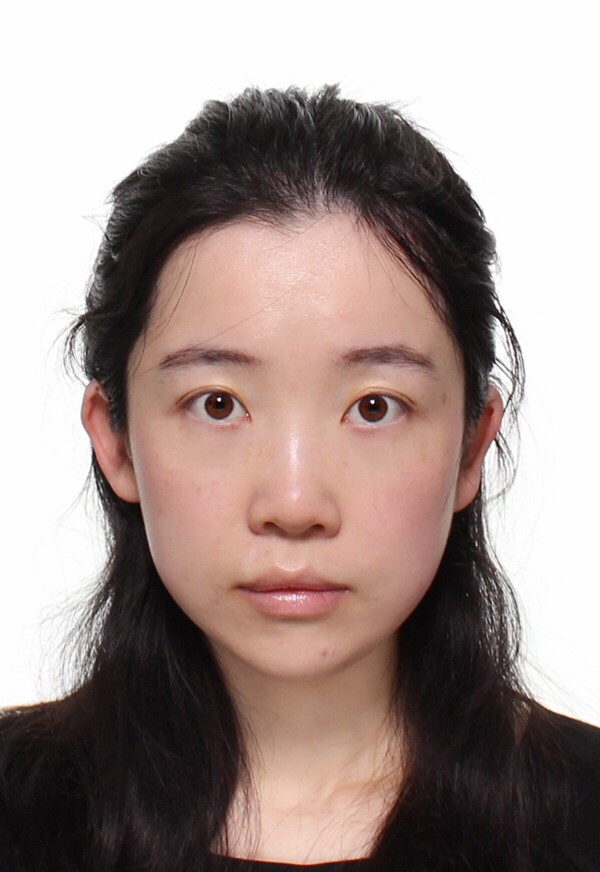}}]{Jingjing Jiang}
(Member, IEEE) received the B.E.
degrees in electrical and electronic engineering from
the University of Birmingham, Birmingham, U.K.,
and the Harbin Institute of Technology, China, in
2010, and the M.Sc. degree in control systems and
the Ph.D. degree from Imperial College London,
London, U.K., in 2011 and 2016, respectively. She
carried out research as part of the Control and Power
Group, Imperial College and joined Loughborough
University as a Lecturer in September 2018. Her re-
search interests include driver assistance control and
autonomous vehicle control design, control design of systems with constraints,
and human-in-the-loop.
\end{IEEEbiography}

\vskip 0pt plus -1fil

\begin{IEEEbiography}[{\includegraphics[width=1in,height=1.25in,clip,keepaspectratio]{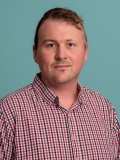}}]{Matthew Coombes}
received the M.Eng. degree in aeronautical engineering and the Ph.D.
degree in autonomous vehicles from the Department of Aeronautical and Automotive Engineering, Loughborough University, Loughbor
ough, U.K., in 2010 and 2015, respectively. He is
currently a Lecturer of Unmanned Vehicles with
Loughborough University, Loughborough, U.K.
He has considerable experience in developing
and operating autonomous platforms, including
system and control design, sensor integration,
and testing. His research interests include autonomous vehicles, some
of which include unmanned aerial system (UAS) forced landing, situational awareness, autonomous taxiing, mission planning optimization
for agricultural aerial surveying and crop spraying, and sensor fusion
and tracking for Counter-UAS. He has recently been awarded with
projects from DASA and Innovate U.K. on Counter UAS and Drone
inspection totaling more than £250 K. Since 2014, he has overseen
all laboratory practical activities for the Loughborough University Centre
for Autonomous Systems.
\end{IEEEbiography}

\vskip 0pt plus -1fil

\begin{IEEEbiography}[{\includegraphics[width=1in,height=1.25in,clip,keepaspectratio]{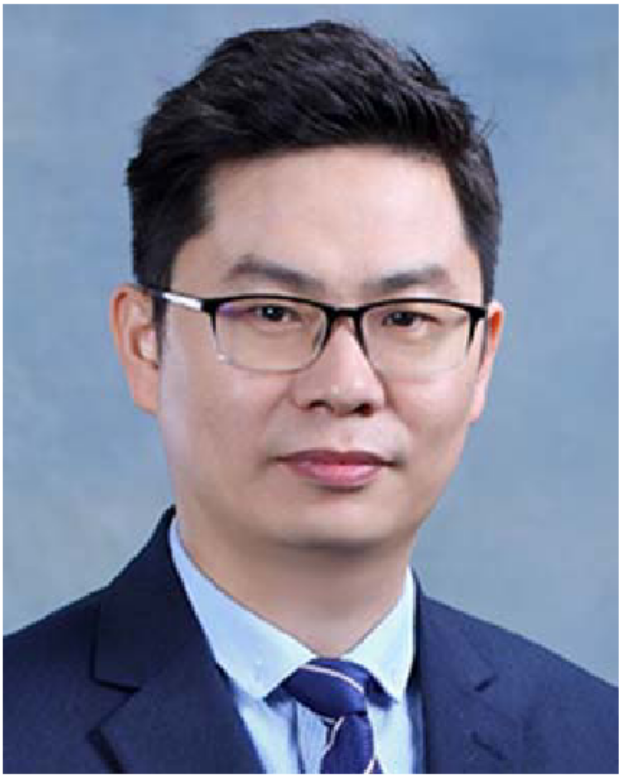}}]{Liang Sun}
(Member IEEE) received the M.Sc. and
Ph.D. degrees in control theory and control engineering from Beihang University, Beijing, China, in 2011 and 2016, respectively. He was a Postdoctoral Fellow of mechanical engineering with Beihang University from 2015 to 2017 and a Visiting Scholar with Loughborouugh University, Loughborough, U.K., from 2019 to 2020. In 2017, he joined the University of Science and Technology Beijing,
Beijing, as an Associate Professor. His research interests includes non-linear mechanical systems control, aerospace control, and intelligent control.
\end{IEEEbiography}




\end{document}